%% file: example_paper.tex
\pdfoutput=1

\documentclass{article}

\usepackage{microtype}
\usepackage{graphicx}
\usepackage{subfigure}
\usepackage{booktabs} 
\usepackage{makecell}
\usepackage{skak}

\usepackage{hyperref}
\usepackage{tikz}
\usetikzlibrary{backgrounds}
\usetikzlibrary{calc}
\usetikzlibrary{fit}
\usetikzlibrary{positioning}

\usepackage[shortcuts]{extdash}

\definecolor{mygreen}{HTML}{167dde}
\definecolor{myred}{HTML}{f22835}

\colorlet{greenfill}{mygreen!20!white}
\colorlet{redfill}{myred!20!white}
\colorlet{moreredfill}{myred!40!white}

\newcommand\loss{L}
\newcommand\hood{\mathcal{S}}

\newcommand{\D}{\mathcal{D}}
\newcommand{\E}{\mathbb{E}}
\newcommand{\R}{\mathbb{R}}
\usepackage{amsmath}

\usepackage{amsfonts}

\renewcommand\bar\overline
\renewcommand\epsilon\varepsilon

\usepackage[accepted]{icml2018}

\icmltitlerunning{Fortified Networks, Under Review at ICML 2018}

\begin{document}

\twocolumn[
\icmltitle{Fortified Networks: Improving the Robustness of Deep Networks\\ by Modeling the Manifold of Hidden Representations}





\begin{icmlauthorlist}
\icmlauthor{Alex Lamb}{}
\icmlauthor{Jonathan Binas}{}
\icmlauthor{Anirudh Goyal}{}
\icmlauthor{Dmitriy Serdyuk}{}
\icmlauthor{Sandeep Subramanian}{}\\
\icmlauthor{Ioannis Mitliagkas}{}
\icmlauthor{Yoshua Bengio}{cifarsenior}
\end{icmlauthorlist}

\centering
Montreal Institute for Learning Algorithms (MILA)

\icmlaffiliation{cifarsenior}{Senior CIFAR Fellow}

\icmlcorrespondingauthor{Alex Lamb}{lambalex@iro.umontreal.ca}

\icmlkeywords{adversarial examples, fortified network}

\vskip 0.3in
]

\begin{abstract}

Deep networks have achieved impressive results across a variety of important tasks. However a known weakness is a failure to perform well when evaluated on data which differ from the training distribution, even if these differences are very small, as is the case with adversarial examples.  We propose \emph{Fortified Networks}, a simple transformation of existing networks, which “fortifies” the hidden layers in a deep network by identifying when the hidden states are off of the data manifold, and maps these hidden states back to parts of the data manifold where the network performs well. Our principal contribution is to show that fortifying these hidden states improves the robustness of deep networks and our experiments (i) demonstrate improved robustness to standard adversarial attacks in both black-box and white-box threat models; (ii) suggest that our improvements are not primarily due to the gradient masking problem and (iii) show the advantage of doing this fortification in the hidden layers instead of the input space.

\end{abstract}

\section{Introduction}

Deep neural networks have been very successful across a variety of tasks. 
This success has also driven applications in domains where reliability and security are critical, including self-driving cars~\citep{bojarski2016end}, health care, face recognition~\citep{sharif2017adversarial}, and the detection of malware~\citep{lecun2015deep}.
Security concerns emerge when an agent using the system could benefit from the system performing poorly.
Reliability concerns come about when the distribution of input data seen during training can differ from the distribution on which the model is evaluated.

\emph{Adversarial examples}~\citep{goodfellow2014adv} is a method to attack neural network models. This attack applies a small perturbation to the input that changes the predicted class. It is notable that it is possible to produce a perturbation small enough that it is not noticeable with a naked eye.
It has been shown that simple gradient methods allow one to find a modification of the input that often changes the output class~\citep{szegedy2013adv,goodfellow2014adv}. More recent work demonstrated that it is possible to create a patch such that even when presented on the camera, it changes the output class with high confidence~\citep{brown2017patch}.

Defences against adversarial examples have been developed as a response. Some of the most prominent classes of defences include feature squeezing~\citep{xu2017squeeze}, adapted encoding of the input~\citep{buckman2018thermometer}, and distillation-related approaches~\citep{papernot2015distill}.  
Existing defenses provide some robustness but most are not easy to deploy. In addition, many have been shown to be vulnerable to gradient masking.  Still others require training a generative model directly in the visible space, which is still difficult today even on relatively simple datasets.  

\begin{figure*}[t]
  \centering
  \includegraphics[width=\textwidth]{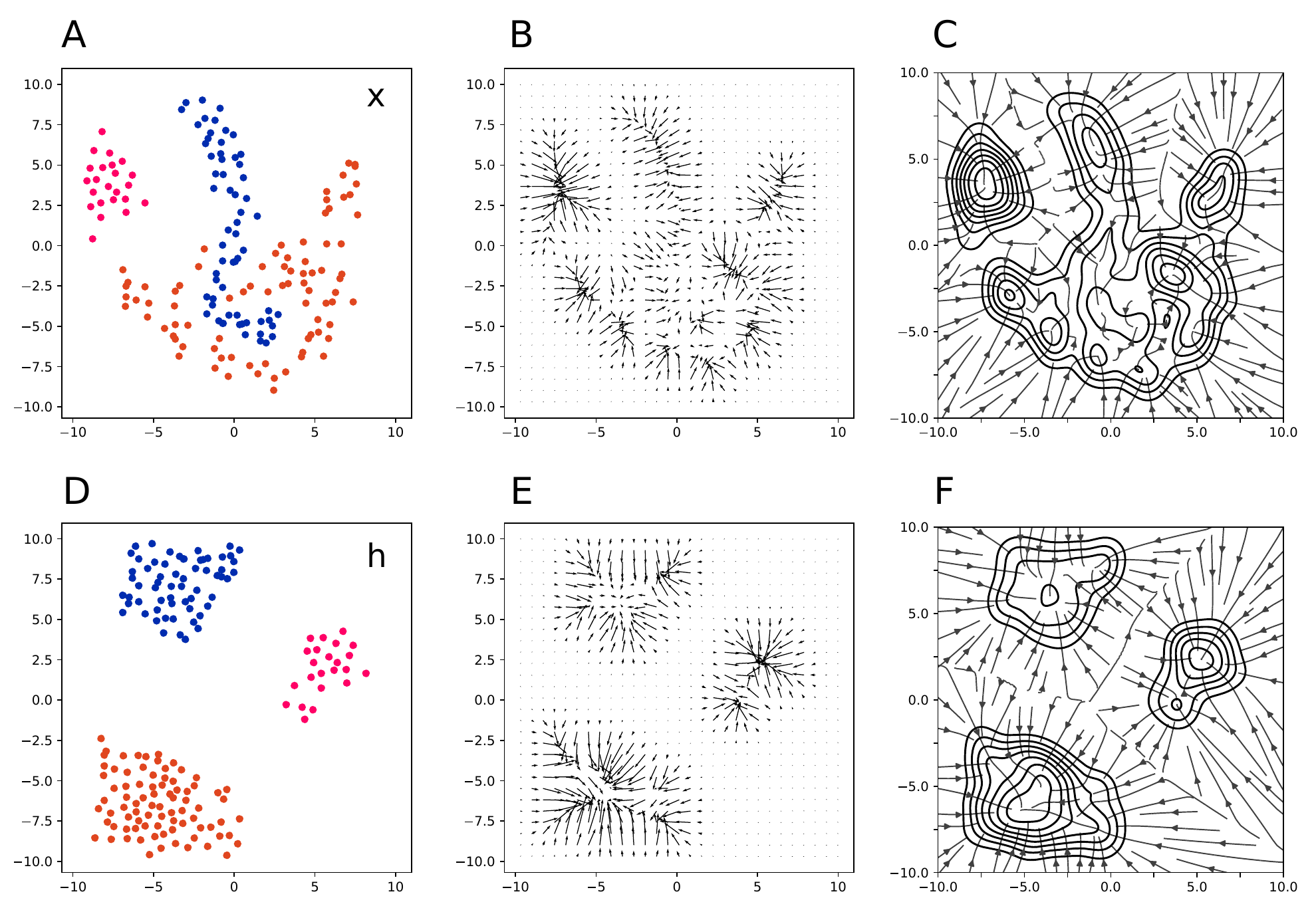}
  \caption{Illustration of the autoencoder dynamics in the input space (top) and in abstract hidden space (bottom).  The leftmost panels show data points from three different classes, the middle panels show vector fields describing the autoencoder dynamics, and the rightmost panels show a number of resulting trajectories and basins of attraction. The key motivation behind Fortified Networks is that directions which point off the data manifold are easier to identify in an abstract space with simpler statistical structure, making it easier to map adversarial examples back to the projected data manifold.  }
  \label{fig:diagram}
\end{figure*}

Our goal is to provide a method which 
(i) can be generically added into an existing network;
(ii) robustifies the network against adversarial attacks and
(iii) provides a reliable signal of the existence of input data that do not lie on the manifold on which it the network trained. 
The ability of generative models, used directly on the input data, to improve robustness is not new.
Our main contribution is that we employ this robustification on the distribution on the learned hidden representations instead making the identification of off-manifold examples easier Figure~\ref{fig:diagram}.

We propose {\em Fortified Networks} \symrook.
Fortification consists of using denoising autoencoders to ``decorate'' the hidden layers of the original network.  We use ``decoration'' in the Pythonic sense that it can be applied to any function (in this case a part of the network) and extend its behavior without explicitly modifying it.  
Fortification meets all three goals stated above.
We discuss the intuition behind the fortification of hidden layers and lay out some of the method's salient properties.
We evaluate our proposed approach on MNIST, Fashion-MNIST, CIFAR10 datasets against whitebox and blackbox attacks.


The rest of the paper is structured in the following way. Section~\ref{sec:background} gives a detailed overview of the background on the adversarial attacks and denoising autoencoders used in this work. Section~\ref{sec:approach} presents our proposed methods for the defence against adversarial examples, Section~\ref{sec:experiments} describes the experimental procedure and Section~\ref{sec:results} provides the experimental results and a comparison to previous approaches. Finally, Section~\ref{sec:related} puts this work into the context of previous publications and Section~\ref{sec:conclusion} concludes.

\section{Background}
\label{sec:background}

\subsection{The Empirical Risk Minimization Framework}

Let us
consider a standard classification task with an underlying data distribution $\D$ 
over pairs of examples $x \in \R^d$ and corresponding labels $y \in [k]$. We also
assume that we are given a suitable loss function $\loss(\theta, x, y)$, for instance the
cross-entropy loss for a neural network. As usual, $\theta \in \R^p$ is the set of
model parameters. Our goal then is to find model parameters $\theta$ that minimize
the risk $\E_{(x, y) \sim \D}[\loss(x, y, \theta)]$. This expectation cannot be computed, therefore a common approach is to to minimize the empirical risk $1/N \sum_D \loss(x, y, \theta)$ taking into account only the examples in a given dataset $D$.

\subsection{Adversarial Attacks and Robustness}

While the empirical risk minimization framework has been very successful and often leads to excellent generalization, it has the significant limitation that it doesn't guarantee robustness, and more specifically performance on examples off the data manifold.  \citet{madry2017adv} proposed an optimization view of adversarial robustness, in which the adversarial robustness of a model is defined as a min-max problem
\begin{align}
\label{eq:minmax}
	\min_\theta \rho(\theta)&,\quad \text{ where }\quad \\
    \rho(\theta) &=
    \mathbb{E}_{(x,y)\sim\mathcal{D}}\left[\max_{\delta\in 
    \hood}
    \loss(\theta,x+\delta,y)\right] \; .
\end{align}

\subsection{Denoising Autoencoders}
\label{sec:daes}


\emph{Denoising autoencoders} (DAEs) are neural networks which take a noisy version of an input (for example, an image) and are trained to predict the noiseless version of that input.  This approach has been widely used for feature learning and generative modeling in deep learning~\citep{Bengio-Courville-Vincent-TPAMI2013}.  More formally, denoising autoencoders are trained to minimize a reconstruction error or negative log-likelihood of generating the clean input. For example, with Gaussian log-likelihood of the clean input given the corrupted input, $r$ the learned denoising function, $C$ a corruption function with Gaussian noise of variance $\sigma^2$, the reconstruction loss is
\begin{equation}
\widehat{\mathcal{L}} = \frac{1}{N} \sum_{n=1}^N
   \left(
      \left\Vert r(C_\sigma(x^{(n)})) - x^{(n)} \right\Vert _2^2
   \right).
   \label{eqn:empirical-loss}
\end{equation}

\citet{alain2012dae} demonstrated that with this loss function, an optimally trained denoising autoencoder's reconstruction vector is proportional to the gradient of the log-density:
\begin{equation}
    \frac{r_{\sigma}(x) - x}{{\sigma^2}} \rightarrow \frac{\partial \log p(x)}{\partial x}
    \hspace{1em} \textrm{as} \hspace{1em} {\sigma} \rightarrow 0. \label{eqn:rx-x-trick}
\end{equation}

This theory establishes that the reconstruction vectors from a well-trained denoising autoencoder form a vector field which points in the direction of the data manifold. However, \citet{alain2012dae} showed that this may not hold for points which are distant from the manifold, as these points are rarely sampled during training. In practice, denoising autoencoders are not just trained with tiny noise but also with large noises, which blurs the data distribution as seen by the learner but makes the network learn a useful vector field even far from the data. 

\section{Fortified Networks}
\label{sec:approach}

\begin{figure*}[t]
    \centering
    \includegraphics{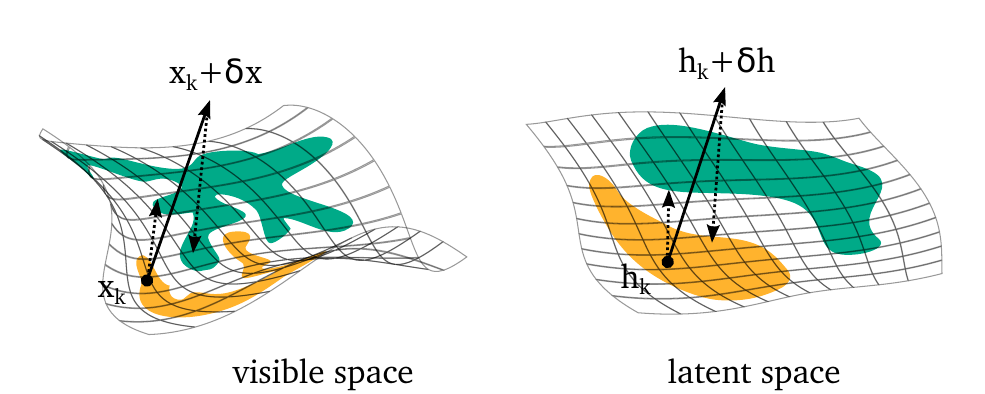}
    \caption{An illustration of the process of mapping back to the manifold in the visible space (left) and the hidden space (right).  The shaded regions represent the areas in the space which are occupied by data points from a given class (they do \emph{not} represent decision boundaries).  }
    \label{fig:manifold3d}
\end{figure*}

\begin{figure*}[t]
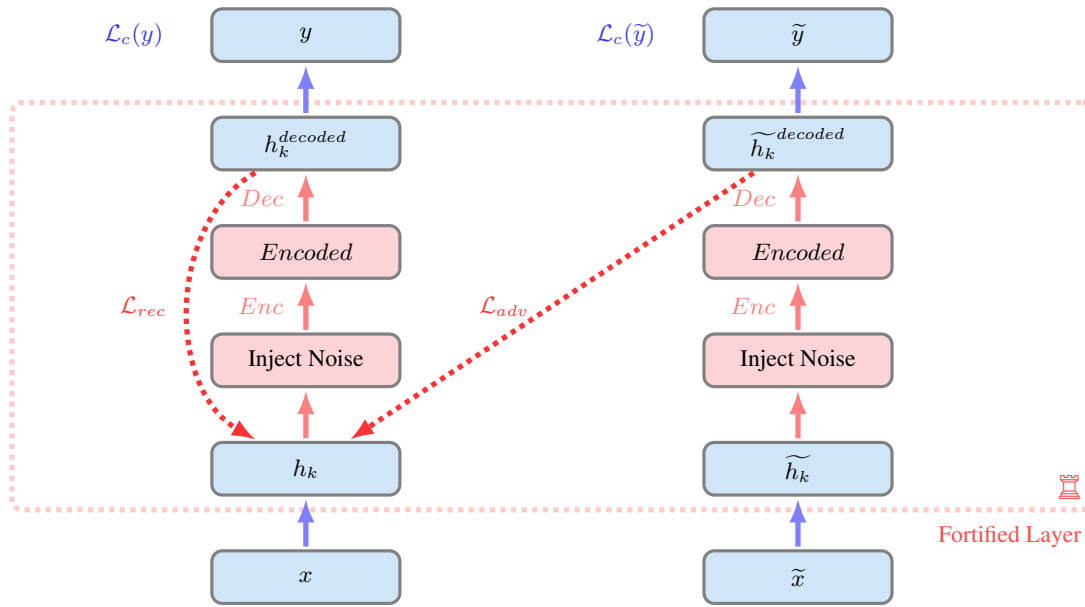

  \centering
  \include{autoencoder}
  \caption{Diagram illustrating a one-layer fortified network. A network is evaluated with a data sample $x$ and its corresponding adversarial example $\widetilde{x}$. Hidden units $h_k$ and $\widetilde{h_k}$ are corrupted with noise, encoded with the encoder $Enc$, and decoded with the decoder $Dec$. The autoencoder (denoted by the red color) is trained to reconstruct the hidden unit $h_k$ that corresponds to the clean input.  Dotted lines are two reconstruction costs: for a benign ($\mathcal{L}_{rec}$) and adversarial examples ($\mathcal{L}_{adv}$). Note that a layer can be fortified at any position in the network.}
  \label{fig:fortified}
\end{figure*}

We propose the use of DAEs inserted at crucial points between layers of the original neural network in order to clean up the transformed data points which may lie away from the original data manifold.
Intuitively, the method aims to regularize the hidden representations by keeping the activations on the surface of the corresponding projected data manifold through the application of a DAE trained on the hidden representations (on the original clean data).
We argue that applying the DAEs on the hidden layers---as opposed to the raw input signal---facilitates learning, while providing a stronger protection from adversarial attacks. As illustrated in Figure~\ref{fig:diagram}, we hypothesize that more abstract representations associated with deeper networks are easier to clean up because the
transformed data manifolds are flatter. The flattening of data manifolds in the deeper layers of a neural network was first noted experimentally by~\citet{Bengio-et-al-ICML2013-small}.
We provide experimental support for these claims in Section~\ref{sec:experiments}.


\paragraph{\symrook \ Layer fortification \symrook} Our method works by substituting a hidden layer $h_k$ with a denoised version.
We feed the signal $h_k$ through the encoder network, $E_k$, and decoder network, $D_k$, of a DAE for layer $k$, which yields the denoised version, $h_k^{decoded}$:
\begin{equation}
    h_k^{decoded} = D_k(E_k(h_k + n_k)),
\end{equation}
where $n_k$ is white Gaussian noise of variance $\sigma^2$ and appropriate shape.
We call the resulting layer, a {\em fortified layer} and the resulting network the {\em fortified network} corresponding to the original network.

For training purposes, we treat the DAEs as part of the fortified network, backpropagate through and train all weights jointly.
Aside from the original classification loss, $\mathcal{L}_c$,
we also include the classification loss from the  adversarial objective, $\mathcal{L}_c(\widetilde{y})$ and 
we introduce a dual objective for the DAEs. 

\begin{itemize}
    \item \textbf{Reconstruction loss.} For a mini-batch of $N$ clean examples, $x^{(1)}, \ldots, x^{(N)}$,
    each hidden layer $h_k^{(1)}, \ldots, h_k^{(N)}$ is fed into a DAE loss, similar to \eqref{eqn:empirical-loss}:
    \begin{equation*}
        \mathcal{L}_{rec,k} = \frac{1}{N} \sum_{n=1}^N
              \left\Vert D_k\left(
                E_k \left( 
                  h_k^{(n)} + n_k
                \right)
              \right)
              - h_k^{(n)} \right\Vert _2^2.
    \end{equation*}
    \item \textbf{Adversarial loss.}
    We use some adversarial training method to produce the perturbed version of the mini-batch,
    $\widetilde{x}^{(1)}, \ldots, \widetilde{x}^{(N)}$, where $\widetilde{x}^{(i)}$ is a small perturbation of $x^{(i)}$ which is designed to make the network produce the wrong answer.
    The corresponding hidden layer $\widetilde{h}_k^{(1)}, \ldots, \widetilde{h}_k^{(N)}$ (using the perturbed rather than original input) is fed into a similar DAE loss:
    \begin{equation*}
        \mathcal{L}_{adv,k} = \frac{1}{N} \sum_{n=1}^N
              \left\Vert D_k\left(
                E_k \left( 
                  \widetilde{h}_k^{(n)} + n_k
                \right)
              \right)
              - h_k^{(n)} \right\Vert _2^2,
    \end{equation*}
    where we note that the target reconstruction for denoising is the clean version of the hidden layer, without noise and without adversarial perturbation.
\end{itemize}

To build a fortified network, we can apply this fortification process to some or all the layers. The final objective used for training the fortified network includes the classification loss and all reconstruction and adversarial losses:
\begin{equation*}
    \mathcal{L} = 
    \mathcal{L}_c(y)+\mathcal{L}_c(\widetilde{y})
    +
    \lambda_{rec} \sum_k \mathcal{L}_{rec,k}
    +
    \lambda_{adv} \sum_k \mathcal{L}_{adv,k},
\end{equation*}
where $ \lambda_{rec}>0$ and  $\lambda_{adv}>0$ tune the strength of the DAE terms. This kind of training process allows for the production of hidden representations robust to small perturbations, and in particular, to adversarial attacks.

\paragraph{Off-manifold signaling} 
The reconstruction losses act as a reliable signal for detecting off-manifold examples (cf. Section~\ref{sec:experiments}). 
This is a particularly useful property in practice: not only can we provide more robust classification results, we can also sense and suggest to the analyst or system when the original example is either adversarial or from a significantly different distribution.

\begin{figure}[t]
  \centering
  \includegraphics[width=\columnwidth]{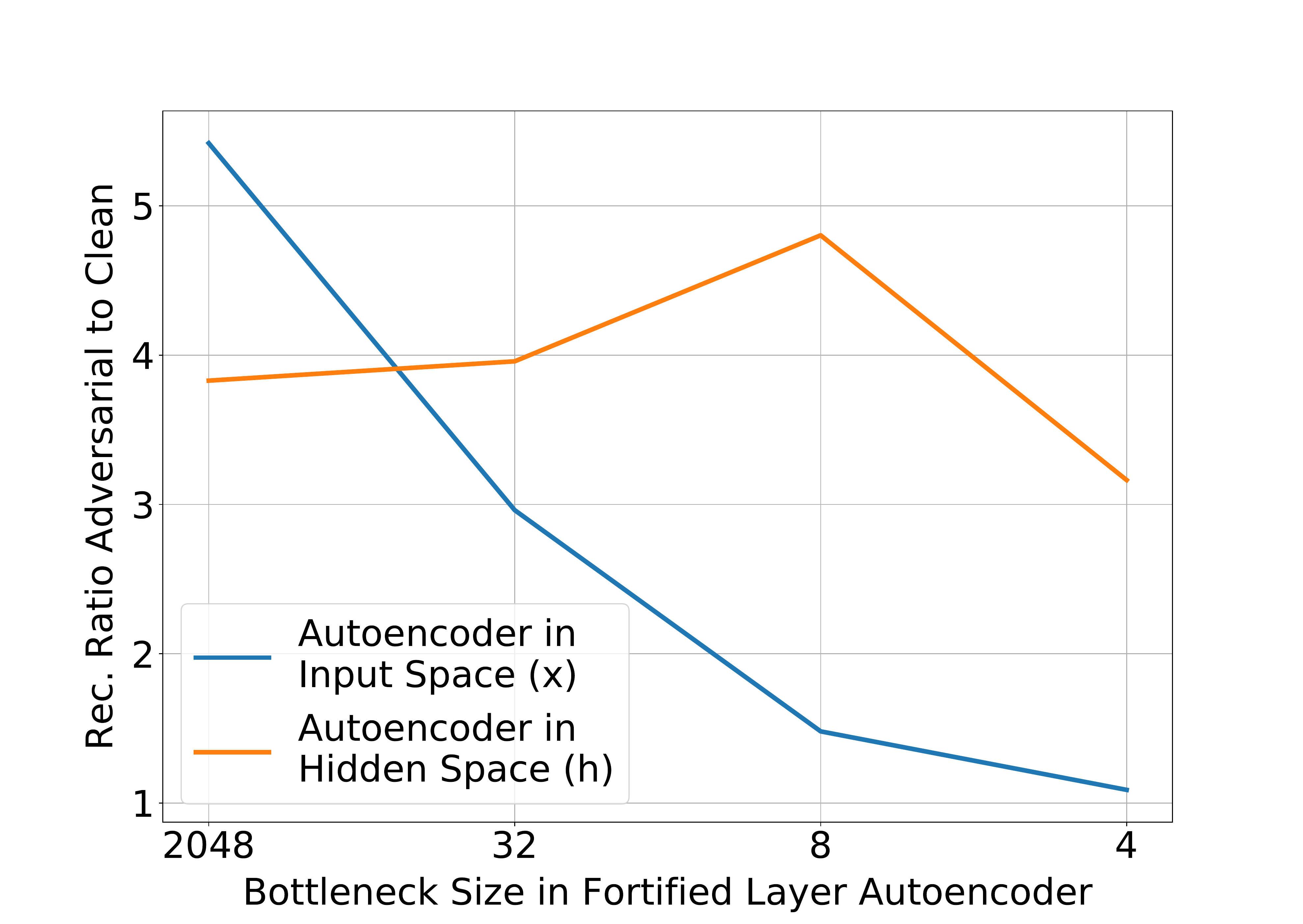}
  \caption{We added fortified layers with different capacities to MLPs trained on MNIST, and display the value of the total reconstruction errors for adversarial examples divided by the total reconstruction errors for clean examples.  A high value indicates that adversarial examples have high reconstruction error.  We considered fortified layers with autoencoders of different capacities.  Our results support the central motivation for fortified networks: that off-manifold points can much more easily be detected in the hidden space (as seen by the relatively constant ratio for the autoencoder in h space) and are much harder to detect in the input space (as seen by this ratio rapidly falling to zero as the autoencoder's capacity is reduced).  }
  \label{fig:rec_error}
\end{figure}

\paragraph{Motivation for when and where to use fortified layers}
We have discussed advantages to placing fortified layers in the hidden states instead of the input space (with further discussion in section ~\ref{sec:gendef}), but the question of where exactly fortified layers need to be placed remains unanswered.  Is it just the final hidden layer?  Is it every hidden layer?  We outline two important considerations regarding this issue:
\begin{enumerate}
    \item In the higher-level hidden layers, it is much easier for the network to identify points which are off of the manifold or close to the margin.  The former is directly experimentally demonstrated in ~\ref{fig:rec_error}.  
    \item At the same time, the higher level hidden layers may already look like points that are not adversarial due to the effect of the adversarial perturbations in the earlier layers.  While we are not aware of any formal study of this phenomenon, it is clearly possible (imagine for example a fortified layer on the output from the softmax, which could only identify unnatural combinations of class probabilities).  
    \item Given these opposing objectives, we argue for the inclusion of multiple fortified layers across the network.  
\end{enumerate}

In the next section we describe a number of experiments to evaluate the practical merit of Fortified Networks.

\section{Experiments}
\label{sec:experiments}

\subsection{Attacks}

We evaluated the performance of our model as a defense against adversarial attacks.  We focused on two of the most popular and well-studied attacks. Firstly, we consider the Fast Gradient Sign Method \citep[FGSM, ][]{goodfellow2014adv} which is popular as it only requires a single step and can still be effective against many networks.  Secondly, we consider the projected gradient descent attack \citep{kurakin2016advml} which is slower than FGSM as it requires many iterations, but has been shown to be a much stronger attack \citep{madry2017adv}.  

Additionally, we consider both white-box attacks (where the attackers knows the model) and black-box attacks (where they don't, but they have access to the training set).  

\subsubsection{Fast Gradient Sign Method}

The Fast Gradient Sign Method (FGSM) ~\cite{goodfellow2014adv} is a simple one-step attack that produces
$\ell_\infty$\=/bounded adversaries via the following gradient based perturbation.
\begin{align}
\widetilde{x} = x + \varepsilon \operatorname{sgn}(\nabla_x \loss(\theta,x,y)).
\end{align}

\subsubsection{Projected Gradient Descent}

The projected gradient descent attack ~\citep{madry2017adv}, sometimes referred to as FGSM$^k$, is a multi-step extension of the FGSM attack characterized as follows:
\begin{align}
x^{t+1} &= \Pi_{x+\hood} \left( x^t +
\alpha\operatorname{sgn}(\nabla_x \loss(\theta,x,y))\right) 
\end{align}
initialized with $x^0$ as the clean input $x$ and with the corrupted input $\widetilde{x}$ as the last step in the sequence.

\subsection{The Gradient Masking and Gradient Obfuscation Problem}

A significant challenge with evaluating defenses against adversarial attacks is that many attacks rely upon a network's gradient.  Methods which reduce the quality of this gradient, either by making it flatter or noisier can lead to methods which lower the effectiveness of gradient-based attacks, but which are not actually robust to adversarial examples \citep{athalye2017robust,papernot2016security}.  This process, which has been referred to as gradient masking or gradient obfuscation, must be analyzed when studying the strength of an adversarial defense.  

One method for studying the extent to which an adversarial defense gives deceptively good results as a result of gradient masking relies on the observation that black-box attacks are a strict subset of white-box attacks, so white-box attacks should always be at least as strong as black-box attacks.  If a method reports much better defense against white-box attacks, it suggests that the selected white-box attack is underpowered as a result of gradient masking.  Another test for gradient masking is to run an iterative search, such as projected gradient descent (PGD) with an unlimited range for a large number of iterations.  If such an attack is not completely successful, it indicates that the model's gradients are not an effective method for searching for adversarial images, and that gradient masking is occurring.  Still another test is to confirm that iterative attacks with small step sizes always outperform single-step attacks with larger step sizes (such as FGSM).  If this is not the case, it may suggest that the iterative attack becomes stuck in regions where optimization using gradients is poor due to gradient masking.    

\subsection{Reconstruction Error as a Heuristic for Deviation from the Manifold}

The theory in \citet{alain2012dae} established that a well-trained denoising autoencoder's reconstruction vector $r(x) - x$ points in the direction of the gradient of the log-density.  Thus, critical points in the model's log-density will have a reconstruction error of zero.  While it is not guaranteed to hold for arbitrary densities, we investigated whether reconstruction error can serve as a practical heuristic for how much a point deviates from the data manifold.  If each data point were a local maximum in the log-density and the log-density has no other critical points, then points on the data manifold are guaranteed to have lower reconstruction error.  

\section{Results}
\label{sec:results}
For details about the specifics of our model architectures and hyperparameters we refer readers to Sections~\ref{sec:appendix-whitebox} and~\ref{sec:appendix-blackbox} of the Appendix. With all experiments, we use the same attacks (with identical parameters) at training and test time to generate adversarial examples.  An important point to note here is that all of the autoencoders in our fortified layers used a single hidden layer with tied weights.  In the case of convolutional autoencoders we always used a stride of 1 and (5,5) kernels.  

{\renewcommand{\arraystretch}{1.2}
\begin{table}[ht]
\centering
\caption{Accuracies against white-box MNIST attacks with FGSM, where the model is a convolutional net.  We used the standard FGSM attack parameters with an $\epsilon$ of 0.3 and compare against published adversarial training defenses.  We also performed ablations where we considered removing the reconstruction error on adversarial examples $\mathcal{L}_{adv}$ as well as switching the activation function in the fortified layers from leaky relu to tanh, which we found to slightly help in this case.  While our baseline and pre-fortified networks used relu activations, we found that by using a leaky relu in all layers the accuracy on FGSM $\epsilon=0.3$ could be improved to 99.2\% with standard adversarial training, suggesting that both our own baselines and those reported in the past have been too weak.  \\}
\label{tb:conv_mnist_whitebox_final_h}
\begin{tabular}{lr} 
\toprule
Model & FGSM  \\ 
\midrule
Adv. Train \citep{madry2017adv} & 95.60  \\ 
Adv. Train \cite{buckman2018thermometer} & 96.17  \\ 
Adv. Train (ours) & 96.36 \\ 
Adv. Train No-Rec (ours) & 96.47 \\ 
\midrule
Quantized \citep{buckman2018thermometer} & 96.29 \\ 
One-Hot \citep{buckman2018thermometer} & 96.22 \\ 
Thermometer \citep{buckman2018thermometer} & 95.84 \\ \hline
\multicolumn{2}{l}{\textit{Our Approaches}} \\ \hline
Fortified Network - Conv, w/o $\mathcal{L}_{adv}$ & 96.46 \\ 
Fortified Network - Conv & \textbf{97.97} \\ 
\bottomrule
\end{tabular}

\end{table}
}

We also ran the above experiment with FGSM and an $\epsilon$ of 0.1 to compare directly with ~\citep{erraqabi2018a3t} and obtain 98.34\% accuracy on adversarial examples compared to their 96.10\%.

\begin{table}[ht]
\centering
\caption{Accuracies against white-box MNIST attacks with PGD with an $\epsilon$ of 0.1, where our model is a convnet.\\}
\label{tb:conv_mnist_whitebox_pgd_final_h}
\begin{tabular}{lr}
\toprule
Model & PGD  \\ 
\midrule
Baseline Adv. Train & 96.98\\ 
Fortified Network - Conv (ours) & \textbf{98.09}\\ 
\bottomrule
\end{tabular}

\end{table}

{\renewcommand{\arraystretch}{1.2}
\begin{table}[ht]
\centering
\caption{Accuracies against white-box CIFAR attacks with FGSM using ($\epsilon=0.3$), where each model is a convnet. Our baseline adversarial training is the resnet model provided in \citep{papernot2017cleverhans}\\ }
\label{tb:conv_mnist_whitebox_final_h}
\begin{tabular}{lr} 
\toprule
Model & FGSM  \\ 
\midrule
Baseline Adv. Train & 79.57  \\ 
Fortified Networks - Conv (ours) & \textbf{80.47}  \\ 
\bottomrule
\end{tabular}
\end{table}
}

{\renewcommand{\arraystretch}{1.2}
\begin{table}[ht]
\centering
\caption{Accuracies against white-box CIFAR attacks with FGSM using the standard ($\epsilon=0.03$), where each model is a convnet. Our baseline adversarial training is the resnet model provided in \citep{papernot2017cleverhans}\\ }
\label{tb:conv_mnist_whitebox_final_h}
\begin{tabular}{lr} 
\toprule
Model & FGSM  \\ 
\midrule
Baseline Adv. Train (ours) & 79.34 \\ 
Quantized (Buckman) & 53.53 \\ 
One-Hot (Buckman) & 68.76 \\ 
Thermometer (Buckman) & 80.97  \\ 
\hline
\multicolumn{2}{l}{\textit{Fortified Networks (ours)}} \\ \hline
Autoencoder on input space \\with loss in hidden states & 79.77 \\
Autoencoder in hidden space & \textbf{81.80}  \\ 
\bottomrule
\end{tabular}
\end{table}
}

{\renewcommand{\arraystretch}{1.2}
\begin{table}[ht]
\centering
\caption{Accuracies against white-box attacks on Fashion MNIST.  For PGD we used $\epsilon=0.1$ and for FGSM we experimented with $\epsilon=0.1$ and $\epsilon=0.3$. Compared with DefenseGAN~\citep{samangouei2018defensegan}. \\}
\label{tb:conv_mnist_whitebox_final_h}
\begin{tabular}{llll} 
\toprule
\thead{Model} & \thead{FGSM \\ ($\epsilon=0.1$)} & \thead{FGSM \\ ($\epsilon=0.3$)} & \thead{PGD \\ ($\epsilon=0.1$)} \\ 
\midrule
DefenseGAN  & n/a & 89.60 & n/a \\ \hline
\multicolumn{4}{l}{\textit{Our Approaches}} \\ \hline
Baseline Adv. Train\\ - Conv,ReLU & 86.14 & 90.66 & 77.49 \\
Baseline Adv. Train\\ - Conv,LReLU & 89.10 & 88.8 & 77.90 \\
Fortified Nets - Conv\\ (ours) & \textbf{89.86} & \textbf{91.31} & \textbf{79.54}  \\
\bottomrule
\end{tabular}
\end{table}
}



{\renewcommand{\arraystretch}{1.2}
\begin{table}[ht]
\centering
\caption{Accuracies against blackbox MNIST attacks with adversarial training. Reporting 50/50 results compared to previous works \citep[JB]{buckman2018thermometer} and \citep[PS]{samangouei2018defensegan}.  The test error on clean examples is in parenthesis.  \\}
\label{tb:conv_mnist_blackbox_final_h}
\begin{tabular}{ll} 
\toprule
Model  & FGSM  \\ 
\midrule
OneHot (JB) &  95.96 (98.83) \\
ThermoEnc (JB) & 96.97 (98.08) \\
DefenseGAN fc$\rightarrow$conv (PS) & 92.21 (n/a) \\
DefenseGAN conv$\rightarrow$conv (PS) & 93.12 (n/a) \\
Adv. Train fc$\rightarrow$conv (PS) & 96.68 (n/a)\\
Adv. Train conv$\rightarrow$conv (PS)  & 96.54 (n/a) \\ \hline
\multicolumn{2}{l}{\textit{Our Approaches}} \\ \hline
Baseline Adv. Train & 93.83 (98.95) \\ 
Fortified Network w/o $\mathcal{L}_{adv}$, $\mathcal{L}_{rec}$ & 96.98 (99.17) \\ 
Fortified Network  & \textbf{97.82} (98.93) \\ 
\bottomrule
\end{tabular}
\end{table}
}

\subsection{Recurrent Networks}

RNNs are often trained using teacher forcing, which refers to the use of the ground-truth samples $y_t$ being fed back into the model and conditioning the prediction of later outputs. These fed back samples force the RNN to stay close to the ground-truth sequence. However, when generating at test time, during the ground truth sequence is not available. We investigated if Fortified Networks could be used to detect when sampling from a teacher-forcing model moves off the manifold.  To this end we train a language model on the standard Text8 dataset, which is derived from Wikipedia articles.  We trained a single-layer LSTM with 1000 units at the character-level, and included fortified layers between the hidden states and the output on each time step.  As seen in table \ref{tb:rnn}, the ratios of these reconstruction errors increases steadily as we increase the number of sampling steps and diverge away from the distribution of training sequences, providing empirical support for the notion that fortified layers effectively measure when the data moves off of the training manifold.

{\renewcommand{\arraystretch}{1.2}
\begin{table}[ht]
\centering
\caption{We trained Fortified Networks on a single-layer LSTM on the Text-8 dataset, with fortified layers added between each step.  We recorded the ratio between reconstruction error on the testing set during both teacher forcing mode and sampling mode (where the model is supplied with its own outputs as inputs for the next step).  The motivation is that the outputs should gradually move off of the manifold with more sampling steps, which is indicated by a high reconstruction error ratio, which makes it an interesting tool for monitoring or potentially fixing this problem.\\}
\label{tb:rnn}
\begin{tabular}{llr} 
\toprule
Sampling Steps & Error Ratio  \\ 
\midrule
50 & 1.03 \\ 
180 & 1.12 \\ 
300 & 1.34 \\ 
\bottomrule
\end{tabular}
\end{table}
}

\begin{figure}[t]
  \centering
  \includegraphics[width=\columnwidth]{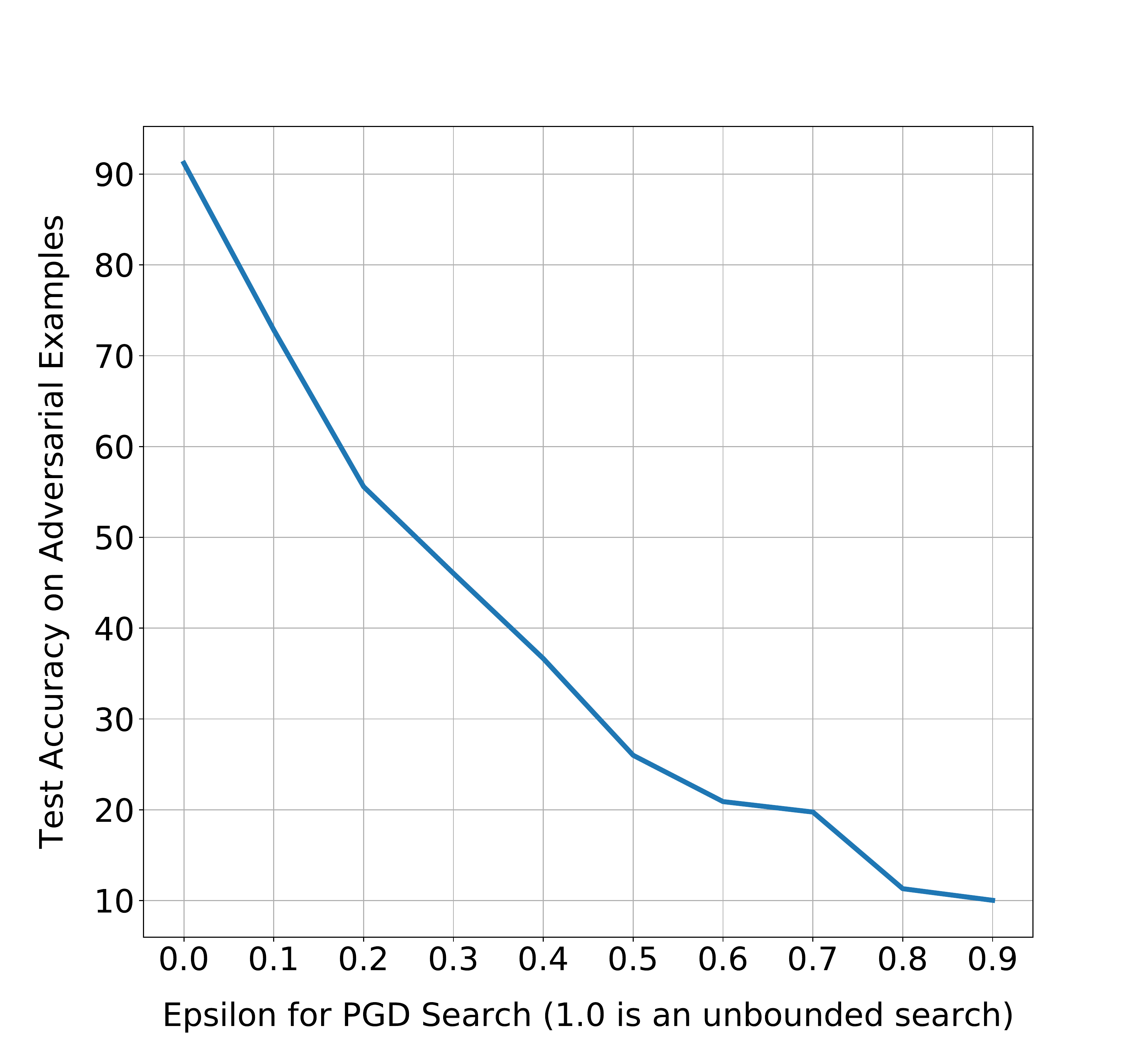}
  \caption{We ran a fortified network on Fashion-MNIST using adversarial training with PGD for a variety of $\epsilon$ values, each for 5 epochs.  The motivation behind this experiment, suggested by \citet{athalye2018obfuscate} is confirming if unbounded ($\epsilon=1$) adversarial attacks are able to succeed.  A defense which succeeds primarily by masking or obfuscating the gradients would fail to bring the accuracy to zero even with an unbounded attack.  As can be seen, unbounded attacks against Fortified Networks succeed when given a sufficiently large $\epsilon$, which is evidence against gradient masking.  }
  \label{fig:fashiion-mnist}
\end{figure}

\section{Related Work}
\label{sec:related}



\subsection{Using Generative Models as a Defense}
\label{sec:gendef}

The observation that adversarial examples often consist of points off of the data manifold and that deep networks may not generalize well to these points motivated ~\citep{gu2014robust,ilyas2017robust,samangouei2018defensegan,liao2017defense} to consider the use of the generative models as a defense against adversarial attacks.  \citet{ilyas2017robust,gilmer2018sphere} also showed the existence of adversarial examples which lie on the data manifold, and \citep{ilyas2017robust} showed that training against adversarial examples forced to lie on the manifold is an effective defense.  Our method shares a closely related motivation to these prior works, with a key difference being that we propose to consider the manifold in the space of learned representations, instead of considering the manifold directly in the visible space.  One motivation for this is that the learned representations have a simpler statistical structure \citep{bengio2012mix}, which makes the task of modeling this manifold and detecting unnatural points much simpler.  Learning the distribution directly in the visible space is still very difficult (even state of the art models fall short of real data on metrics like Inception Score) and requires a high capacity model.  Additionally working in the space of learned representations allows for the use of a relatively simple generative model, in our case a small denoising autoencoder.  

\citet{ilyas2017robust} proposed to work around these challenges from working in the visible space by using the Deep Image Prior instead of an actual generative model.  While this has the advantage of being a model that doesn't require a special training procedure (as deep image prior is a separate optimization process for each example) it may be limited in the types of adversarial attacks that it's resistant to, and it would provide no defense against adversarial attacks which are in the range of a convolutional network, which have been shown to exist \citep{xiao2018generating}.  

Another key difference between our work and \citep{ilyas2017robust,samangouei2018defensegan} is that both DefenseGAN and the Invert-and-Classify approach use an iterative search procedure at inference time to map observed data points onto nearby points on the range of the generator.  On the other hand, our approach uses small denoising autoencoders that are used in the same way (i.e. a simple forward application) during both training and testing.  The use of such an iterative procedure presents challenges for evaluation, as it is possible for gradients to vanish while doing backpropagation through such a procedure, which may lead to an overestimate in the strength of the defense due to the gradient masking problem \citep{papernot2016blackbox,athalye2018obfuscate}.  One indicator of the gradient masking problem is black-box attacks outperforming white-box attacks, which is an indicator of under-powered attacks as black-box attacks are a strict subset of white-box attacks.  This indicator of gradient obfuscation was present in the work of \citet{samangouei2018defensegan} where black-box attacks were generally stronger against their defense, but with our method we observe very similar defense quality against black-box and white-box attacks.  

\citep{gu2014robust,liao2017defense} both considered using an autoencoder as a pre-processing step in the input space.  Interestingly \citep{liao2017defense} used a loss function defined in the space of the hidden states, but still used autoencoders directly in the input space.  

\subsection{Adversarial Hidden State Matching}

\citet{erraqabi2018a3t} demonstrate that adversarially matching the hidden layer activations of regular and adversarial examples improves robustness.  This work shared the same motivation of using the hidden states to improve robustness, but differed in that they used an adversarial objective and worked in the original hidden states instead of using a generative model (in our case, the DAE in the fortified layers).  We present direct experimental comparisons with their work in section \ref{sec:results}.  

\subsection{Denoising Feature Matching}

\citet{warde2016improving} proposed to train a denoising autoencoder in the hidden states of the discriminator in a generative adversarial network.  The generator's parameters are then trained to make the reconstruction error of this autoencoder small.  This has the effect of encouraging the generator to produce points which are easy for the model to reconstruct, which will include true data points.  Both this and Fortified Networks use a learned denoising autoencoder in the hidden states of a network.  A major difference is that the denoising feature matching work focused on generative adversarial networks and tried to minimize reconstruction error through a learned generator network, whereas our approach targets the adversarial examples problem.  Additionally, our objective encourages the output of the DAE to denoise adversarial examples so as to point back to the hidden state of the original example, which is different from the objective in the denoising feature matching work, which encouraged reconstruction error to be low on states from samples from the generator network.  

\subsection{Adversarial Spheres}

\citet{gilmer2018sphere} studied the existence of adversarial examples in the task of classifying between two hollow concentric shells.  Intriguingly, they prove and construct adversarial examples which lie on the data manifold (although \citet{ilyas2017robust} also looked for such examples experimentally using GANs).  The existence of such on-manifold adversarial examples demonstrates that a simplified version of our model trained with only $\mathcal{L}_{rec}$ could not protect against all adversarial examples.  However, training with $\mathcal{L}_{adv}$ encourages the fortified layers to map back from points which are not only off of the manifold, but also to map back from points which are hard to classify, allowing Fortified Networks to also potentially help with on-manifold adversarial examples as well.

\section{Conclusion}
\label{sec:conclusion}

Protecting against adversarial examples could be of paramount importance in mission-critical applications.
We have presented Fortified Networks, a simple method for the robustification of existing deep neural networks.
Our method is
\begin{itemize}
    \item Practical:
fortifying an existing network entails introducing DAEs between the hidden layers of the network and can be automated.
We are preparing a PyTorch module that does exactly that and will release it for the deep learning community to use shortly. Furthermore, the DAE reconstruction error at test time is a reliable signal of distribution shift, that is examples unlike those encountered during training. 
High error can signify either adversarial attacks or significant domain shift; both are important cases for the analyst or system to be aware of.
    \item Effective:
    We showed results that improve upon the state of the art on defenses for adversarial attacks on MNIST and provides improvement on CIFAR and Fashion-MNIST.  
\end{itemize}


\paragraph{Limitations}
The cost of the proposed method is the extended training time due to the search for an adversarial example and training the autoencoder.  The added cost of the fortified layers over adversarial training by itself is relatively small, and is also much easier and simpler than training a full generative model (such as a GAN) in the input space.  Layer fortification typically involves smaller DAEs that require less computation.  Additionally, we have shown improvements on ResNets where only two fortified layers are added, and thus the change to the computational cost is very slightly.  At the same time, fortified networks have only been shown to improve robustness when used alongside adversarial training, which is expensive for iterative attacks.

\bibliography{example_paper}
\bibliographystyle{icml2018}

\clearpage

\input{appendix.tex}



\end{document}

%% file: autoencoder.tex
\begin{tikzpicture}[
    black!50, text=black,
    font=\small,
    node distance=4mm,
    dnode/.style={
        align=center,
        rectangle,minimum size=10mm,rounded corners,
        inner sep=5pt},
    rnode/.style={
        align=center,
        rectangle,
        minimum width=25mm,
        minimum height=7mm,
        rounded corners,
        inner sep=3pt,
        very thick,draw=black!50},
    tuplenode/.style={
        align=center,
        rectangle,minimum size=10mm,rounded corners,
        inner sep=5pt},
    darrow/.style={
        rounded corners,-latex,shorten <=5pt,shorten >=1pt,line width=2mm},
    mega thick/.style={line width=2pt}]
    
\matrix[row sep=7mm,column sep=20mm] {
    \node (y) [rnode,left color = greenfill, right color = greenfill] {$y$}; &&
      \node (y_tilde) [rnode,left color = greenfill, right color = greenfill] {$\widetilde{y}$};\\
    \node (h) [rnode,left color = greenfill, right color = greenfill] {$h_k^{decoded}$}; &&
      \node (h_tilde) [rnode,left color = greenfill, right color = greenfill] {$\widetilde{h_k}^{decoded}$};\\
    \node (e) [rnode,right color=redfill, left color = redfill] {$Encoded$};&&
      \node (e_tilde) [rnode,right color=redfill, left color = redfill] {$Encoded$};\\
    \node (n) [rnode,right color=redfill, left color = redfill] {Inject Noise};&&
      \node (n_tilde) [rnode,right color=redfill, left color = redfill] {Inject Noise};\\
    \node (hd) [rnode,left color = greenfill, right color = greenfill] {$h_k$}; &&
      \node (hd_tilde) [rnode,left color = greenfill, right color = greenfill] {$\widetilde{h_k}$};\\
    \node (x) [rnode,left color = greenfill, right color = greenfill] {$x$};&&
      \node (x_tilde) [rnode,left color = greenfill, right color = greenfill] {$\widetilde{x}$};\\
};
\draw[-latex,shorten <=1pt,shorten >=1pt,mega thick,color=red!50]  (e_tilde) to node[label=left:$Dec$]{} (h_tilde);
\draw[-latex,shorten <=1pt,shorten >=1pt,mega thick,color=red!50]  (hd_tilde) to (n_tilde);
\draw[-latex,shorten <=1pt,shorten >=1pt,mega thick,color=red!50]  (n_tilde) to node[label=left:$Enc$]{} (e_tilde);
\draw[-latex,shorten <=1pt,shorten >=1pt,mega thick,color=blue!50] (x_tilde) to (hd_tilde);
\draw[-latex,shorten <=1pt,shorten >=1pt,mega thick,color=blue!50] (h_tilde) to (y_tilde);

\draw[-latex,shorten <=1pt,shorten >=1pt,mega thick,color=red!50]  (e) to node[label=left:$Dec$]{} (h);
\draw[-latex,shorten <=1pt,shorten >=1pt,mega thick,color=red!50]  (hd) to (n);
\draw[-latex,shorten <=1pt,shorten >=1pt,mega thick,color=red!50]  (n) to node[label=left:$Enc$]{} (e);
\draw[-latex,shorten <=1pt,shorten >=1pt,mega thick,color=blue!50] (x) to (hd);
\draw[-latex,shorten <=1pt,shorten >=1pt,mega thick,color=blue!50] (h) to (y);

\draw[-latex,shorten <=1pt,shorten >=1pt,mega thick,color=red!80,dotted] (h_tilde)   to node[label=left:$\mathcal{L}_{adv}$]{} (hd);
\draw[-latex,shorten <=1pt,shorten >=1pt,mega thick,color=red!80,dotted] (h)  to [bend right = 60] node[label=left:$\mathcal{L}_{rec}$]{}  (hd) ;


\node (yloss) [left=0.5cm of y, color=blue!80] {$\mathcal{L}_{c}(y)$};
\node (yloss_tilde) [left=0.5cm of y_tilde,blue!80] {$\mathcal{L}_{c}(\widetilde{y})$};

\begin{pgfonlayer}{background}
    \node (cycle) [fit={(h_tilde) (hd) (e) ($(h_tilde.east)+(70pt,0pt)$) ($(hd.west)+(-70pt,0pt)$)},rnode,inner sep=5pt,draw=red!20,mega thick,dotted] {};
\end{pgfonlayer}
\node [below left=2pt of cycle.south east,text=red!70] {\footnotesize Fortified Layer};
\node [above left=2pt of cycle.south east,text=red!70] { \large \symrook};

\end{tikzpicture}

%% file: appendix.tex
\appendix
\section*{Appendix}

\section{Experimental Setup}
All attacks used in this work were carried out using the Cleverhans ~\citep{papernot2017cleverhans} library.
\subsection{White-box attacks}
\label{sec:appendix-whitebox}
Our convolutional models (Conv, in the tables) have 2 strided convolutional layers with 64 and 128 filters followed by an unstrided conv layer with 128 filters. We use ReLU activations between layers then followed by a single fully connected layer. The convolutional and fully-connected DAEs have a single bottleneck layer with leaky ReLU activations with some ablations presented in the table below.

With white-box PGD attacks, we used only convolutional DAEs at the first and last conv layers with Gaussian noise of $\sigma=0.01$ whereas with FGSM attacks we used a DAE only at the last fully connected layer. The weight on the reconstruction error $\lambda_{rec}$ and adversarial cost $\lambda_{adv}$ were set to 0.01 in all white-box attack experiments. We used the Adam optimizer with a learning rate of 0.001 to train all models. 

The table below lists results a few ablations with different activation functions in the autoencoder

{\renewcommand{\arraystretch}{1.2}
\begin{table}[ht]
\centering
\caption{More detailed version of table \ref{tb:conv_mnist_whitebox_final_h}, but with more detailed ablation experiments for our method included.  Accuracies against white-box MNIST attacks with FGSM, where the model is a convolutional net.  We used the standard FGSM attack parameters with an $\epsilon$ of 0.3 and compare against published adversarial training defenses.  We also performed ablations where we considered removing the reconstruction error on adversarial examples $\mathcal{L}_{adv}$ as well as switching the activation function in the fortified layers from leaky relu to tanh, which we found to slightly help in this case. \\}
\label{tb:conv_mnist_whitebox_final_h_appendix}
\begin{tabular}{lr} 
\toprule
Model & FGSM  \\ 
\midrule
Adv. Train \citep{madry2017adv} & 95.60  \\ 
Adv. Train \citep{buckman2018thermometer} & 96.17  \\ 
Adv. Train (ours) & 96.36 \\ 
Adv. Train No-Rec (ours) & 96.47 \\ 
\midrule
Quantized \citep{buckman2018thermometer} & 96.29 \\ 
One-Hot \citep{buckman2018thermometer} & 96.22 \\ 
Thermometer \citep{buckman2018thermometer} & 95.84 \\ \hline
\multicolumn{2}{l}{\textit{Our Approaches}} \\ \hline
Fortified Network - Conv, w/o $\mathcal{L}_{adv}$, lrelu & 96.46 \\ 
Fortified Network - Conv, lrelu  & 97.69 \\ 
Fortified Network - Conv, tanh & \textbf{97.97} \\ 
\bottomrule
\end{tabular}

\end{table}
}

\subsection{Black-box attacks}
\label{sec:appendix-blackbox}
Our black-box results are based on a fully-connected substitute model (input-200-200-output), which was subsequently used to attack a fortified convolutional network.
The CNN was trained for 50 epochs using adversarial training, and the predictions of the trained CNN were used to train the substitute model.
6 iterations of Jacobian data augmentation were run during training of the substitute, with $\lambda=0.1$. The test set data holdout for the adversary was fixed to 150 examples.
The learning rate was set to 0.003 and the Adam optimizer was used to train both models.



